\documentclass[11pt,a4paper]{article}
\usepackage[hyperref]{format/emnlp2020}
\usepackage{times}
\usepackage{graphicx}
\usepackage{latexsym}
\usepackage{booktabs}
\usepackage{color}
\usepackage{subfig}
\usepackage{multirow}
\usepackage{amssymb}
\usepackage{amsmath}
\usepackage{algorithm}
\usepackage{algorithmic}
\usepackage{bm}

\usepackage{microtype}

\aclfinalcopy 

\title{
The Devil is the Classifier:  \\
Investigating  Long Tail Relation Classification with Decoupling Analysis}

\author{
Haiyang Yu$^{1,2}$\thanks{\quad Equal contribution and shared co-first authorship.}, 
Ningyu Zhang$^{1,2*}$,
Shumin Deng$^{3}$, 
Zonggang Yuan$^{3}$, \\
\textbf{Yantao Jia}$^{1,2}$, 
\textbf{Huajun Chen}$^{1,2}$ \thanks{\quad Corresponding author: C.Hua(huajunsir@zju.edu.cn)} \\
$^{1}$ Zhejiang University \\
$^{2}$ AZFT Joint Lab of Knowledge Engine \\
$^{3}$  Huawei Technologies Co., Ltd \\
 {\tt \{yuhaiyang,zhangningyu,231sm,huajunsir\}@zju.edu.cn}\\
 {\tt jamaths.h@163.com yuanzonggang@huawei.com} 
}

\date{}

\begin{document}
\maketitle

\begin{abstract}
Long-tailed relation classification is a challenging problem as the head classes may dominate the training phase, thereby leading to the deterioration of the tail performance. Existing solutions usually address this issue via class-balancing strategies, e.g., data re-sampling and loss re-weighting, but all these methods adhere to the schema of entangling learning of the \emph{representation} and \emph{classifier}.  In this study, we conduct an in-depth empirical investigation into the long-tailed problem and found that \textbf{pre-trained models with instance-balanced sampling already capture the well-learned representations for all classes}; moreover, it is possible to achieve better long-tailed classification ability at low cost by \textbf{only adjusting the classifier}. Inspired by this observation, we propose a robust classifier with \emph{attentive relation routing}, which assigns soft weights by automatically aggregating the relations. Extensive experiments on two datasets demonstrate the effectiveness of our proposed approach. Code and datasets are available in \url{https://github.com/zjunlp/deepke}.
\end{abstract}

\section{Introduction}
Relation classification (RC) is a fundamental task in natural language processing (NLP) for constructing knowledge graphs and downstream-related tasks, such as question answering \cite{Jin2019ComQAQA} and information retrieval \cite{zamani2020generating}. Most approaches for relation extraction \cite{zhang2018capsule,nan2020reasoning,zhang2020relation,zhang2021,ye2021} primarily focus on frequently seen relations. However, in more practical scenarios, large numbers of samples with long-tailed distributions of the relation frequencies are inevitable. Expressly, nearly  \textbf{70\%} of the relations (NA excluded) in the widely used TACRED \cite{zhang2017position} dataset have long-tailed characteristics. Therefore, it is crucial for the models to be able to extract such long-tailed relations. 

For the long-tailed data, a commonly noted issue is that the head classes dominate the training stage, thereby leading to the deterioration of performance on the tail. To address this issue, one feasible solution is to learn a better \textbf{representation}~\cite{zhang2019balance} via re-sampling strategies or transfer learning for long-tailed relations. Another solution is to design specific loss functions that better facilitate learning with long-tailed data to acquire a robust \textbf{classifier} for the tail. 

Recently proposed methods \cite{wang2017learning,zhang2019long} have generally studied the problem of extracting long-tailed relations via jointly learning representation and classifier.  However, the mechanisms behind such jointly schema are not thoroughly understood, thus making it unclear how the long-tailed classification ability is achieved--\emph{is it from learning a better representation or by handling the discriminability better via robust classifier decision boundaries?} Note that pre-trained language models such as BERT \cite{devlin2018bert:} have the ability to represent various texts robustly; hence, we can argue whether a well-performing classifier is more critical for the long-tailed problem.  To answer the above questions, in this study, we conducted an empirical investigation on long-tailed RC. 

To this end, we first conduct experiments (\S~\ref{prob}) to identify the working mechanism of representation and classifier learning. We observe that previous re-weighting operations achieve performance gains, while the re-sampling approaches slightly deteriorate the performance when retrained with fixed encoder; this necessitated that \textbf{classifier is more important than representation learning for the long-tailed problem}. Hence, we further introduce an effective classifier, namely \emph{attention relation routing}. We argue that previous approaches \cite{Joulin2016LearningVF,lin2017focal,cui2019class} assigns hard weights, which are suboptimal; therefore, we propose a soft weighting mechanism for classifier training with dynamic routing \cite{sabour2017dynamic} (\S~\ref{capsule}). Experimental results on Fewrel-LT and TACRED  demonstrate the efficacy of our proposed approach.  Our key contributions can be summarized as follows:

\begin{itemize}
\item  We empirically investigate long-tailed relation extraction and observe that \textbf{classifier} is more important than representation learning.
\item  We introduce \emph{attentive relation routing} to assign soft weights for strong classifier learning.
\item  We report the empirical results on two datasets and release a long-tailed RC testbed. 
\end{itemize}

\section{Background}
There are two main methods by which the long-tailed issue can be addressed \cite{Buda2018ASS}. The first involves \emph{re-balancing strategies}, which are intuitive methods to solve the problem of long-tailed distributions. Concerning these, \citet{Joulin2016LearningVF} proposed the class-balanced sampling (CBS) approach to sample instances for each class. Other sampling methods, such as square-root sampling (SRS)~\cite{mahajan2018exploring}, progressively balanced sampling (PBS)~\cite{Kang2020DecouplingRA}, mixed class and instance distributions, are also introduced. The second type of solution involves \emph{re-weighting methods}, including re-weight loss (RWL) \cite{Ronneberger2015UNetCN}, focal loss (FCL) \cite{lin2017focal}, and dice loss (DSL) \cite{Milletari2016VNetFC}, which  assign  weights  to different training samples for each class to boost the discriminability via robust classifier decision boundaries. More recently, ~\cite{Zhou2019BBNBN} proposed a unified bilateral-branch network (BBN) to comprehensively consider both representation learning and classifier learning for long-tailed recognition. In \cite{Kang2020DecouplingRA}, the authors introduced a $\tau$-norm approach to re-balance the classifiers' decision boundaries. Besides, \cite{wang2017learning,zhang2019long} propose to transfer knowledge from -head to -tail. However, most of the approaches train the classifiers entangled with the representations, thus rendering the essence of the mode abilities not to be understood adequately.

\section{Empirical Decoupling Analysis} \label{prob}

\begin{figure}
\centering
\includegraphics[width=0.36\paperwidth]{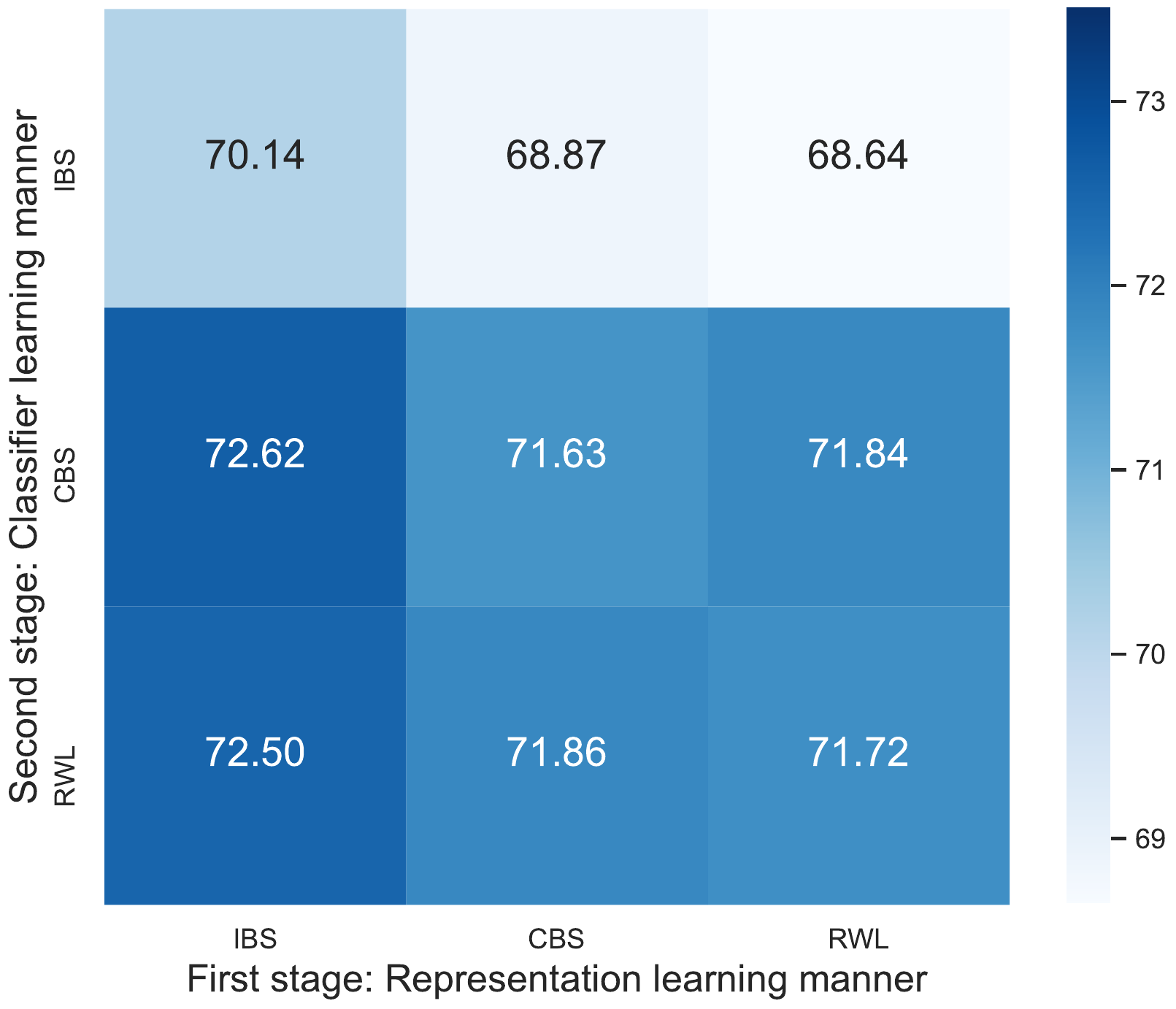}
\caption{Macro F1  on FewRel-LT dataset with two-stage empirical decoupling analysis.  We observed 1) F1 score of classifiers trained with CBS/RWL are much \emph{higher} than IBS (vertical direction), which indicates that \textbf{adjusting only the classifier benefit a lot}; 2) F1 score trained with CBS/RWL surprisingly \emph{deteriorate}  the performance (horizontal direction), which implies that \textbf{CBS/RWL lead to inferior discriminative ability of the learned deep features.}}
\label{fig:compare-rc}
\end{figure}

In this section, we present to investigate the working mechanism for representation learning and classifier training by empirical decoupling analysis. We divide the model into two parts: the feature extractor (i.e., backbone networks such as BERT) and the classifier (i.e., last fully connected layers). 


\subsection{Representation \& Classifier}
Since BERT~\cite{devlin2018bert:} has been proved to be powerful and effective in many NLP tasks, we first preprocess the sentence  $\mathbf{x}=$ $\{w_1,$ $w_2,$ $e_1,$ $\dots,$ $e_2$ $w_n\}$ to BERT's input form: $\mathbf{x}=$ $\{$[CLS]$,$ $w_1,$ $w_2,$ [E1], $e_1,$ [/E1], $\dots,$ [E2], $e_2,$ [/E2], $w_n,$ [SEP]$\}$, where $w_i, i \in  [1, n]$ refers to each word in a sentence; $e_1$ and $e_2$ are two labeled entities; [E1], [/E1], [E2], and [/E2] are four special tokens used to mark the positions of the entities.  Note that we are aimed to investigate the representation and classifier, we utilize the simplest way, namely, the [CLS] token,  as the sentence feature representation: $\mathbf{x} = [CLS]$, $x \in \mathbb{R}^{d_{bert}}$, and $d_{bert}$ is the predefined output size in BERT. 

Generally, the BERT output is followed by a fully connected layer to match sample features with relation categories that need to be classified. Other kinds of classifiers can also be leveraged. We utilize the softmax function to predict relation logits as follows:

\begin{equation}
    \mathbf{y}_{pred} = softmax(\mathbf{W}\mathbf{x} + b)
\end{equation}
 
\subsection{Decoupling Procedure}
To decouple the representation learning and classifier training, we conduct a two-stage training process on   the FewRel-LT dataset\footnote{The construction process of the FewRel-LT dataset is described in \S~\ref{exp}.}, \emph{In the first stage}, we train three models with instance-balanced sampling (vanilla setting, \textbf{IBS}), class-balanced sampling (\textbf{CBS}), and  re-weighted  loss (\textbf{RWL})  strategies. \emph{In the second stage}, we fix the parameters of the feature extractors and retrain the classifiers \emph{from scratch} with the learning mentioned above methods. In principle, we design these experiments to fairly compare the quality of the representations and classifiers learned via different schema by following the control variates method.

From Figure \ref{fig:compare-rc}, we empirically observe that 1) when we apply the same representation learning technique, it is observed that the RWL/CBS always achieve better performance than the IBS, which is attributable to their re-balancing operations adjusting the classifier weights and updating them to match the test distributions;  2) when applying the same classifier learning scheme, it can be seen that the F1 score of IBS is consistently higher than that of RWL/CBS. The worst results of the RWL/CBS method reveal that they lead to inferior discriminability of the deep learned features. These empirical observations indicate that the previous performance gains are mainly based on \text{enhancing the discriminability}, which indicates that a robust classifier is essential for the long-tailed problem. 

\section{Classifier for Long-Tailed RC} \label{capsule}
Motivated by the observations mentioned above, we believe that a more effective classifier can boost performance. Note that the previous approaches (e.g., focal loss \cite{lin2017focal} and dice loss \cite{Milletari2016VNetFC}) assign hard weights to adjust and update the classifier to match the test distributions. Inspired by the dynamic routing scheme \citet{sabour2017dynamic,tsai2020capsules}, which is able to assign soft weights via route-by-agreement, we introduce a simple yet effective approach, namely  \textbf{attentive relation routing (ARR}), to learn the classifier weights.

\subsection{Attentive Relation Routing}
 Specifically,  given a sentence representation $\mathbf{x}$, we transform it into the primary capsules ($\bm{\Omega}^{p}$), where $\bm{\Omega}^{p}$ $\in$ $\mathbb{R}^{a \times m}$, ${a}$ is the number of primary capsules, and ${m}$ is the hidden size of the capsule. The overall routing can be summarized as\footnote{More details are provided in the supplementary materials} follows: 
\begin{equation}
    \mathbf{y}_{pred} = \textbf{Attentive\_Routing}(\mathbf{x} \mapsto \bm{\Omega}^{p})
\end{equation}

The routing procedure consists of multiple iterations, and each iteration process consists of two steps. The first step computes the agreement coefficient between the primary capsules ($\bm{\Omega}^p$) and relation capsules ($\bm{\Omega}^r$), and the second step updates the relation capsules by the calculated routing coefficient. Here, $\bm{\Omega}^r$ $\in$ $\mathbb{R}^{b \times n}$, $b$ is the number of relation capsules, and $n$ is the capsule dimension.

We first transform $\bm{\Omega}^p$ to the \textbf{vote} ($\mathcal{V}$) using a learned transformation matrix: $\mathcal{V} = \mathsf{W} \cdot \bm{\Omega}^p$,
where the matrix $\mathsf{W}$ $\in$ $\mathbb{R}^{b \times a}$.
Next, the \textbf{agreement} (${\alpha}_{mn}$) is computed by the dot-product similarity between the relation capsules and vote:
${\alpha}_{mn} = {\bm{\Omega}^r}^{\top} \cdot \mathcal{V}$,
After that we can get routng coefficient: $r_{mn}=softmax({\alpha}_{mn})$.
Inspired by \cite{tsai2020capsules}, we updated $\bm{\Omega}^r$ with layer normalization rather than the squash function as follows:
$\bm{\Omega}^r = LayerNorm(\sum r_{mn} \mathcal{V})$.
After several routing iterations, we use a single linear transformation to squeeze each relation capsule into a scalar and  generate the final prediction result  as follows:
$\mathbf{y}_{pred} = softmax(\mathsf{W}_r \bm{\Omega}^r + b)$,
where $\mathsf{W}_r$ $\in$ $\mathbb{R}^{n \times 1}$.

\section{Experiments}

\begin{table}
\centering
\begin{tabular}{l|c|ccc}
\toprule
Dataset & class & train & valid & test \\
\midrule
FewRel-LT & 50&5,499 &5,000&5,000\\

TACRED & 42 & 68,124&22,631&15,509\\
\bottomrule
\end{tabular}
\caption{Statistics of long-tailed datasets.}
\label{tab:dataset}
\end{table}

\subsection{Datasets \& Settings}\label{exp}
We performed experiments on two long-tailed relation classification datasets: FewRel~\cite{han2018fewrel:} and TACRED~\cite{zhang2017position}. 
For TACRED, we used the original dataset. However, for FewRel, we randomly selected 50 classes and reconstructed a long-tailed dataset, FewRel-LT, by reducing the number of training samples per class according to an exponential function $n_i=n_{max}\eta^i$, where $i$ is the class index, $n_{max}$ are the training samples in the largest class (index 0), and $\eta$ is the imbalance ratio defined as dividing the maximum number of the class by the minimum. We take $n_{max}$ equal to 500,  $\eta$ equal to 100.
For the validation and test datasets, we select  100 samples for each class.
Additional dataset statistics are shown in Table \ref{tab:dataset}.

We utilize \emph{bert-base-uncased} as the representation from \citet{Wolf2019HuggingFacesTS}. We employed Adam~\cite{Kingma2015AdamAM} as the optimizer, the initial learning rate $\alpha$ is set to 0.01 (the learning rate of the classifier, more details in supplementary materials), and we reduce the rate by 20\% every 8 epochs.
The coefficient for the regularization term $\lambda$ is 5e-4, the batch size is 64, and the total number of epochs is 50.
We evaluate the performance of FewRel-LT with \emph{macro F1} score and TACRED with \emph{offical micro F1} score.

\subsection{Main Results}

\begin{table}
\centering
\resizebox{0.48\textwidth}{!}{%
\begin{tabular}{l|c|c}
\toprule
Dataset & \small{FewRel-LT} & \small{TACRED} \\
\midrule
BERT (instance-balanced sampling)              & 70.14 & 65.2 \\
\midrule
CBS~\cite{Joulin2016LearningVF}         & 71.63 & 65.7 \\
SRS~\cite{mahajan2018exploring}         & 71.80 & 66.3 \\
PBS~\cite{Kang2020DecouplingRA}         & 72.01 & 66.4 \\
\midrule
RWL~\cite{Ronneberger2015UNetCN}        & 71.72 & 65.9 \\
FCL~\cite{lin2017focal}                 & 71.64 & 66.2 \\
DSL~\cite{Li2019DiceLF}                 & 72.48 & 66.8 \\
\midrule
BBN~\cite{Zhou2019BBNBN}                 & 73.17 & 67.5 \\
$\tau$-norm~\cite{Kang2020DecouplingRA}  & 72.93 & 67.3 \\
\midrule
Our ARR(one iteration)         & 74.92 & 68.2 \\
Our ARR(two iteration)         & \textbf{75.30} & 68.5 \\
Our ARR(three iteration)       & 75.26 & \textbf{68.8} \\
\bottomrule
\end{tabular}
}
\caption{Evaluation results on test data with macro F1 score($10^{-2}$). The first row refers to the vanilla BERT-base approach; CBS, SRS, PBS in the second row refer to different data re-sampling methods; RWL, FCL, DSL in the third row indicates different loss re-weighting strategies; BBN and $\tau$-norm in the fourth row refer to recent state-of-the-art approaches for the long-tailed problem. The last row is our method of attentive relation routing.}
\label{tab:result}
\end{table}

We compare our attentive relation routing (ARR) with extensive baselines (We do not compare with the transfer learning baseline \cite{zhang2019long} as it utilize external class hierarchy.).
From Table \ref{tab:result}, we observe that 
,  data re-sampling strategies such as CBS \cite{Joulin2016LearningVF} and SRS \cite{mahajan2018exploring} obtain  about 1.5\% improvement compared with vanilla BERT;  meanwhile, PBS \cite{Kang2020DecouplingRA} which smoothly transitioned from IBS to CBS, achieves about 2\% improvement. We also observe that re-weighting methods, including RWL \cite{Ronneberger2015UNetCN}, FCL \cite{lin2017focal}, and DSL \cite{Milletari2016VNetFC}, all achieve better performance compared  with vanilla BERT. Further, we find that  BBN~\cite{Zhou2019BBNBN} and $\tau$-norm~\cite{Kang2020DecouplingRA}  both achieve better improvements. Our ARR  achieve \textbf{5\%} and \textbf{3\%} improvements on each dataset.

\subsection{Further Analysis}

\begin{figure}
\centering
\includegraphics[width=0.36\paperwidth]{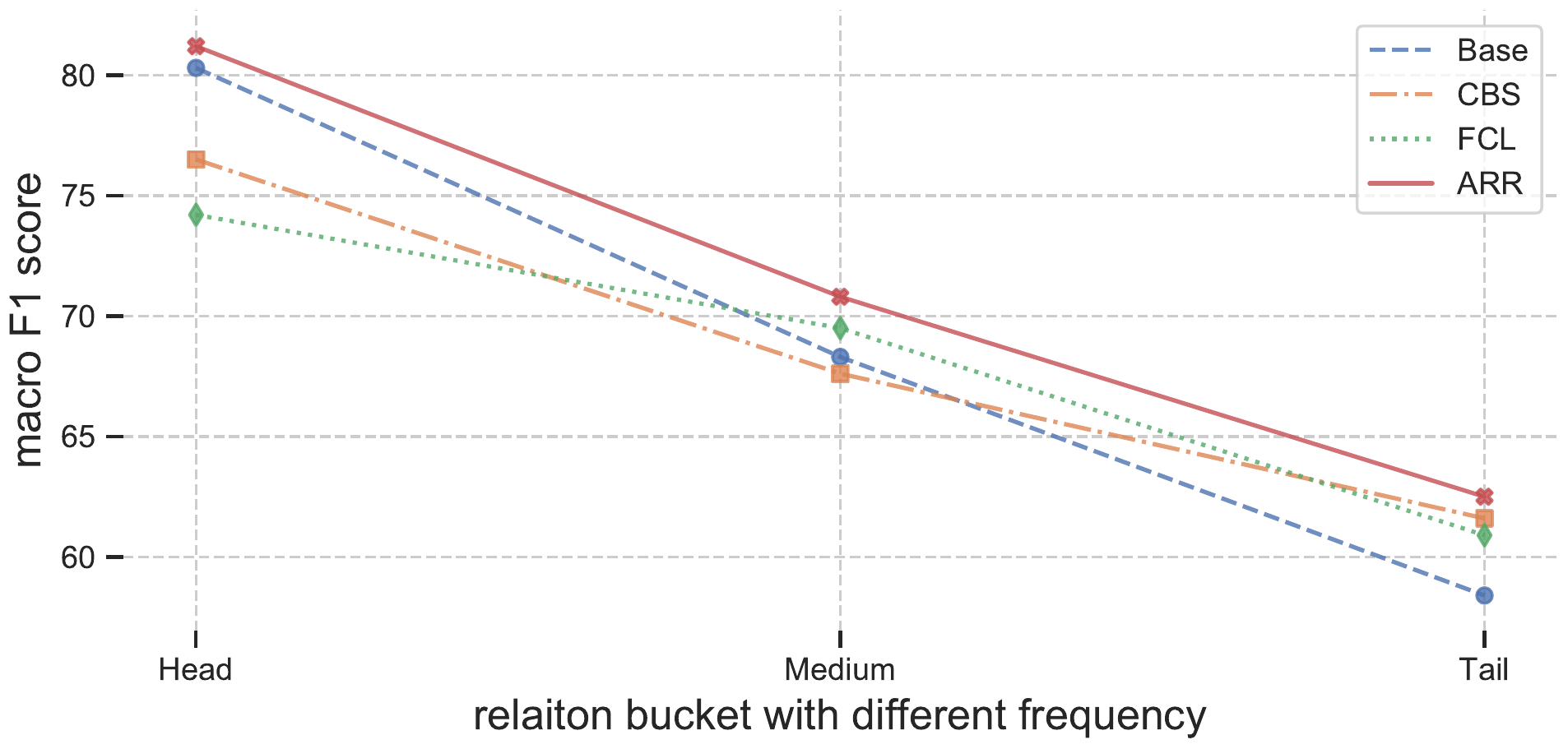}
\caption{Performance analysis with different samples.}
\label{fig:case-study}
\end{figure}

To better analyze the performance variations across classes with different numbers of samples, we further report the \emph{F1} score on subsets of the FewRel-LT: \emph{Head} (more than 100 samples), \emph{Medium} (20$\sim$100 samples), and \emph{Tail} (less than 20 samples).
To prevent the leakage of entity information, we exclude samples with \emph{shared entity pairs}.
From Figure~\ref{fig:case-study}, we observe that the data re-sampling and loss re-weighting methods improve the tail performance, but deteriorate the performance of the head. However, our ARR method improves the tail score without reducing head performance, which indicates that our approach with dynamic weights is able to enhance the discriminability of the tail without reducing the robustness of the head. 
\section{Conclusion and Future Work}
We study the problem of long-tailed relation classification and develop a preliminary step towards decoupling representation and classifier learning. We empirically observed that with pre-trained language models, \emph{a good classifier was the most important requirement} for long-tailed classification, which might shed light on future works with long-tailed  problems. 
We further introduce a robust classifier with attentive relation routing.  Extensive experiments on two benchmark datasets demonstrate the effectiveness of our approach. We anticipate further research on promising directions, including 1) exploiting a more effective classifier to boost long-tailed discriminability, 2) distinguishing task-specific representation, and classifier automatically via neural architecture search.

\bibliography{format/anthology,myref}
\bibliographystyle{format/acl_natbib}



\clearpage

\end{document}


\maketitle

\section{Datasets Details}
\begin{figure}[H]
\centering
\includegraphics[width=0.36\paperwidth]{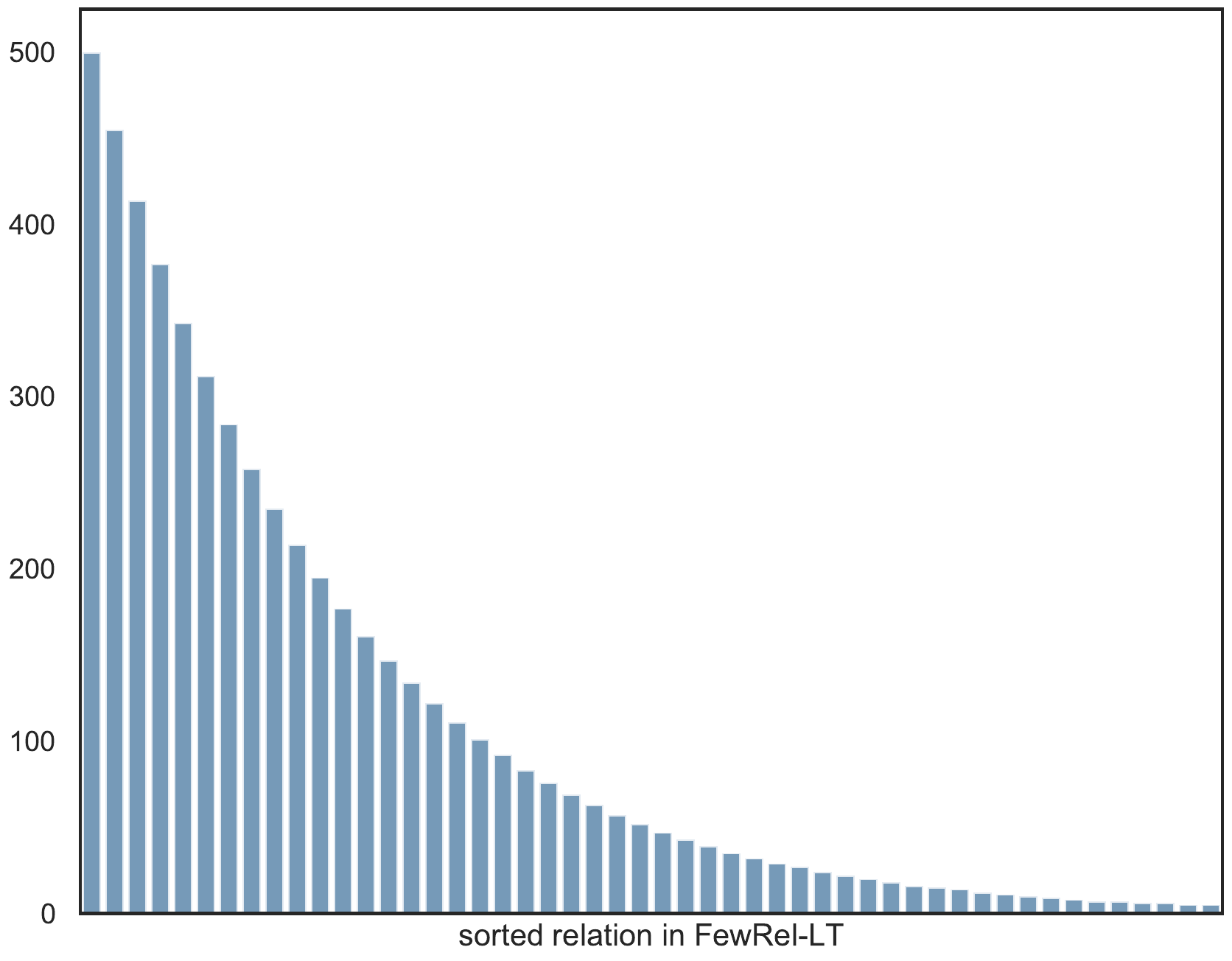}
\caption{Long-tail distribution in Fewrel-LT.}
\label{fewrel-lt}
\end{figure}

\begin{figure}[H]
\centering
\includegraphics[width=0.36\paperwidth]{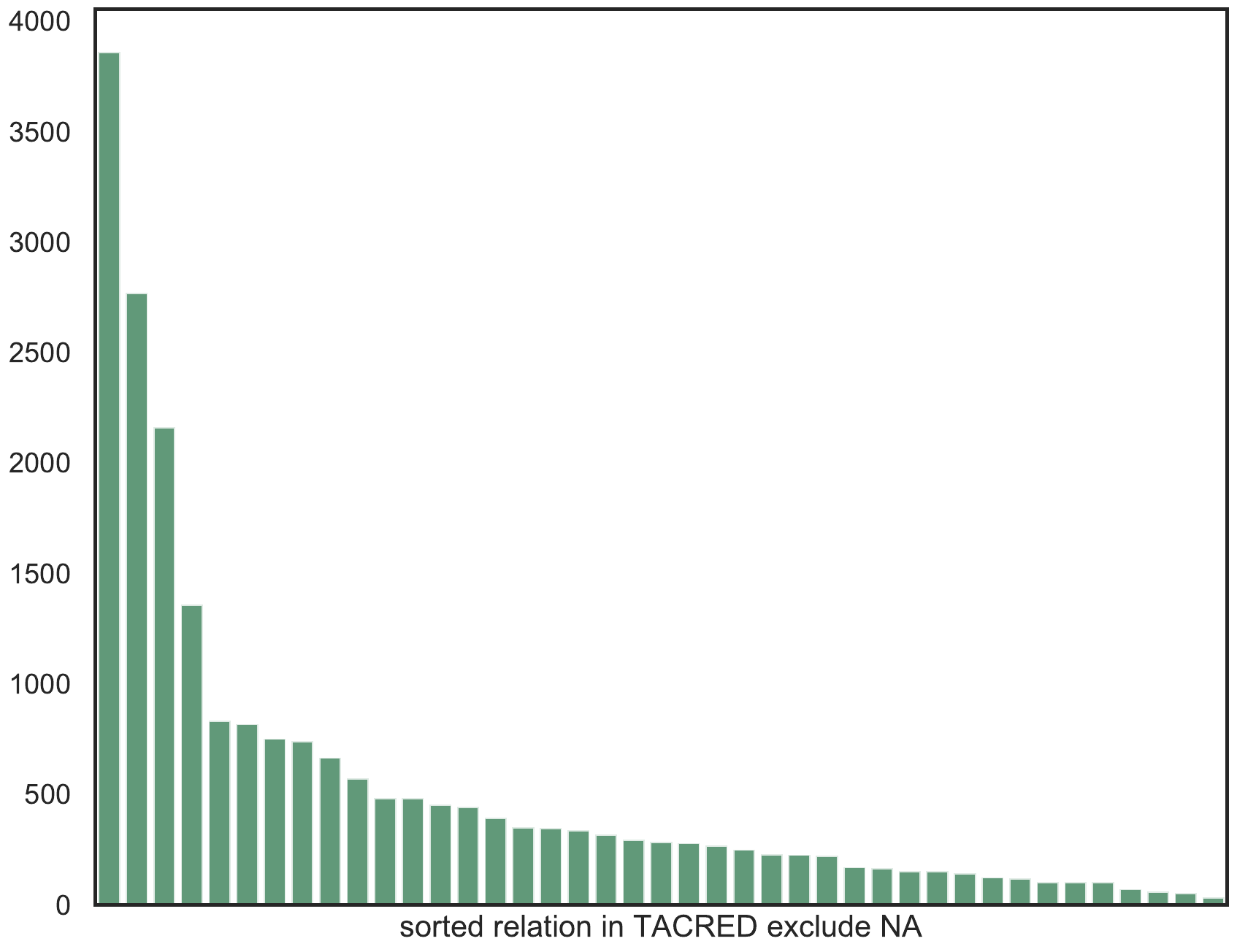}
\caption{Long-tail distribution in TACRED.}
\label{tacred}
\end{figure}

\textbf{Our datasets are available in the supplementary materials for reproducibility.}.
As the Figure \ref{fewrel-lt} and \ref{tacred} shows,  there exist amounts of long-tailed relations  in Fewrel-LT and  TACRED dataset.

\section{Experiments Details}
\textbf{Our code is available in the supplementary materials for reproducibility, the code can be used as a long-tailed RC testbed.}. We detail the training procedures and hyperparameters for each of the datasets. We utilize Pytorch \cite{paszke2019pytorch} to conduct experiments with  Nvidia 1080ti GPU.    All optimization was performed with the Adam optimizer with a linear warmup of learning rate over the first 10\% of gradient updates to a maximum value, then linear decay over the remainder of the training. Gradients were clipped if their norm exceeded 1.0, and weight decay on all non-bias parameters was set to 5e-4. Grid search was used for hyperparameter tuning (maximum values bolded below), using five random restarts for each hyperparameter setting for all datasets. Early stopping was performed on the development set. The batch size was 64 in all cases.

\textbf{Fewrel-LT.}
The Fewrel-LT dataset is available in the supplementary materials. Each epoch cost nearly half  hour. The hyper-parameter search space was:
\begin{itemize}
\item batch size: [8, 16, 32, \textbf{64}]
\item bert learning rate: [1e-3, \textbf{1e-4}, 1e-5]
\item learning rate: [1e-1, 5e-2, \textbf{1e-2}, 5e-3, 1e-3]
\item weight decay: [1e-1, 5e-2, 1e-2, 5e-3, 1e-3, \textbf{5e-4}, 1e-4]
\item primary capsule numbers: [16, 24, 32, \textbf{64}]
\item routing iteration: [1, \textbf{2}, 3]
\end{itemize}

\textbf{TACRED.}
The TACRED dataset is available in url\footnote{https://nlp.stanford.edu/projects/tacred/}. Each epoch cost nearly one hour. The hyper-parameter search space was:
\begin{itemize}
\item batch size: [8, 16, 32, \textbf{64}]
\item bert learning rate: [1e-3, \textbf{1e-4}, 1e-5]
\item learning rate: [1e-1, 5e-2, \textbf{1e-2}, 5e-3, 1e-3]
\item weight decay: [1e-1, 5e-2, 1e-2, 5e-3, 1e-3, \textbf{5e-4}, 1e-4]
\item primary capsule numbers: [16, 24, \textbf{32}, 64]
\item routing iteration: [1, 2, \textbf{3}]
\end{itemize}

Experimental results on the development set are shown in Table~\ref{dev}.

\begin{table}[H]
\centering
\resizebox{0.48\textwidth}{!}{%
\begin{tabular}{l|c|c}
\toprule
Dataset & \small{FewRel-LT} & \small{TACRED} \\
 
\midrule
Our ARR(dev)        & 75.41 & 68.9  \\
Our ARR(test)       & 75.30 & 68.8  \\
\bottomrule
\end{tabular}
}
\caption{Evaluation results on development set.}
\label{dev}
\end{table}

\section{Pseudocode of Attentive Relation Routing}
\begin{algorithm*}[ht]
\caption{Attentive Relation Routing algorithm  between layer $L$ and $L+1$.}\label{algo:routingalg}
\begin{algorithmic}[1]
\Procedure{Attentive Relation Routing}{${\bf P}^{L}$, ${\bf P}^{L+1}$, ${\bf W}^L$}
\State for all capsule $i$ in layer $L$ and capsule $j$ in layer $(L+1)$: ${\bf v}_{ij}^L \gets {\bf W}_{ij}^L\cdot{\bf p}_i^{L}$ \Comment{vote}
\State for all capsule $i$ in layer $L$ and capsule $j$ in layer $(L+1)$: $a_{ij}^L \gets {{\bf p}_{j}^{L+1}}^\top\cdot {\bf v}_{ij}^L$ \Comment{agreement}
\State for all capsule $i$ in layer $L$: $r_{ij}^L \gets \mathrm{exp}(a_{ij}^L) \,/ \sum_{j'} \mathrm{exp}(a_{ij'}^L)$ \Comment{routing coefficient}
\State for all capsule $j$ in layer ($L+1$): ${\bf p}_j^{L+1} \gets \sum_i{r_{ij}^L{\bf  v}_{ij}^L}$ \Comment{pose update}
\State for all capsule $j$ in layer ($L+1$): ${\bf p}_j^{L+1} \gets \texttt{LayerNorm}({\bf p}_j^{L+1})$ \Comment{normalization}
\State\Return ${\bf P}^{L+1}$
\EndProcedure
\end{algorithmic}
\label{algo:routing}
\end{algorithm*}


The overall \emph{attentive relation routing} process can be seen in the algorithm ~\ref{algo:routing}.



\bibliography{format/anthology,app}
\bibliographystyle{format/acl_natbib}